%% file: main.tex
\documentclass[11pt,letterpaper]{article}
\usepackage{emnlp2016}
\usepackage{times}
\usepackage{latexsym}

\usepackage{mystuff-nlp}
\usepackage{url}
\usepackage{color}
\usepackage{float}
\makeatletter
\newcommand{\@BIBLABEL}{\@emptybiblabel}
\newcommand{\@emptybiblabel}[1]{}
\makeatother
\usepackage[draft,hidelinks]{hyperref}
\usepackage{amsfonts,amssymb,amsmath,bm}
\usepackage{algorithm}
\usepackage{appendix}
\usepackage{algpseudocode}
\usepackage{graphicx,subfigure}
\usepackage{booktabs}
\usepackage{multirow}

\usepackage{enumitem}

\newcommand{\mb}[1]{\mathbf{#1}}

\let\argmax\relax

\DeclareMathOperator*{\argmax}{arg\,max}
\newcommand{\notleftright}{\mathrel{\ooalign{$\leftrightarrow$\cr\hidewidth$/$\hidewidth}}}
\newcommand{\sysname}{\textsc{Ntel}}

\newcommand{\example}[1]{`#1'}
\usepackage[table]{xcolor}

\emnlpfinalcopy

\def\emnlppaperid{1053}

\title{Toward Socially-Infused Information Extraction:\\
Embedding Authors, Mentions, and Entities}

\author{Yi Yang\\
	    Georgia Institute of Technology\\
        Atlanta, GA 30308, USA\\
	    {\tt yiyang@gatech.edu}\And
        Ming-Wei Chang\\
        Microsoft Research\\
        Redmond, WA 98052, USA\\
        {\tt minchang@microsoft.com}\And
        Jacob Eisenstein\\
	    Georgia Institute of Technology\\
        Atlanta, GA 30308, USA\\
        {\tt jacobe@gatech.edu}
}

\date{}

\begin{document}
\maketitle
\begin{abstract}
\input{abstract}
\end{abstract}

\input{intro}

\input{data}

\input{homophily}

\input{model}

\input{exp}

\input{related}

\input{con}

\input{ack}

\bibliographystyle{emnlp2016}
\bibliography{cite-strings,cites,cite-definitions}

\clearpage
\input{appendix}

\end{document}

%% file: abstract.tex
Entity linking is the task of identifying mentions of entities in text, and linking them to entries in a knowledge base. This task is especially difficult in microblogs, as there is little additional text to provide disambiguating context; rather, authors rely on an implicit common ground of shared knowledge with their readers. In this paper, we attempt to capture some of this implicit context by exploiting the social network structure in microblogs. We build on the theory of \emph{homophily}, which implies that socially linked individuals share interests, and are therefore likely to mention the same sorts of entities. We implement this idea by encoding authors, mentions, and entities in a continuous vector space, which is constructed so that socially-connected authors have similar vector representations. These vectors are incorporated into a neural structured prediction model, which captures structural constraints that are inherent in the entity linking task. Together, these design decisions yield F1 improvements of 1\%-5\% on benchmark datasets, as compared to the previous state-of-the-art.

%% file: intro.tex
\section{Introduction}
\label{sec:intro}

\begin{figure}[t!]
\centering
\includegraphics[scale=.22]{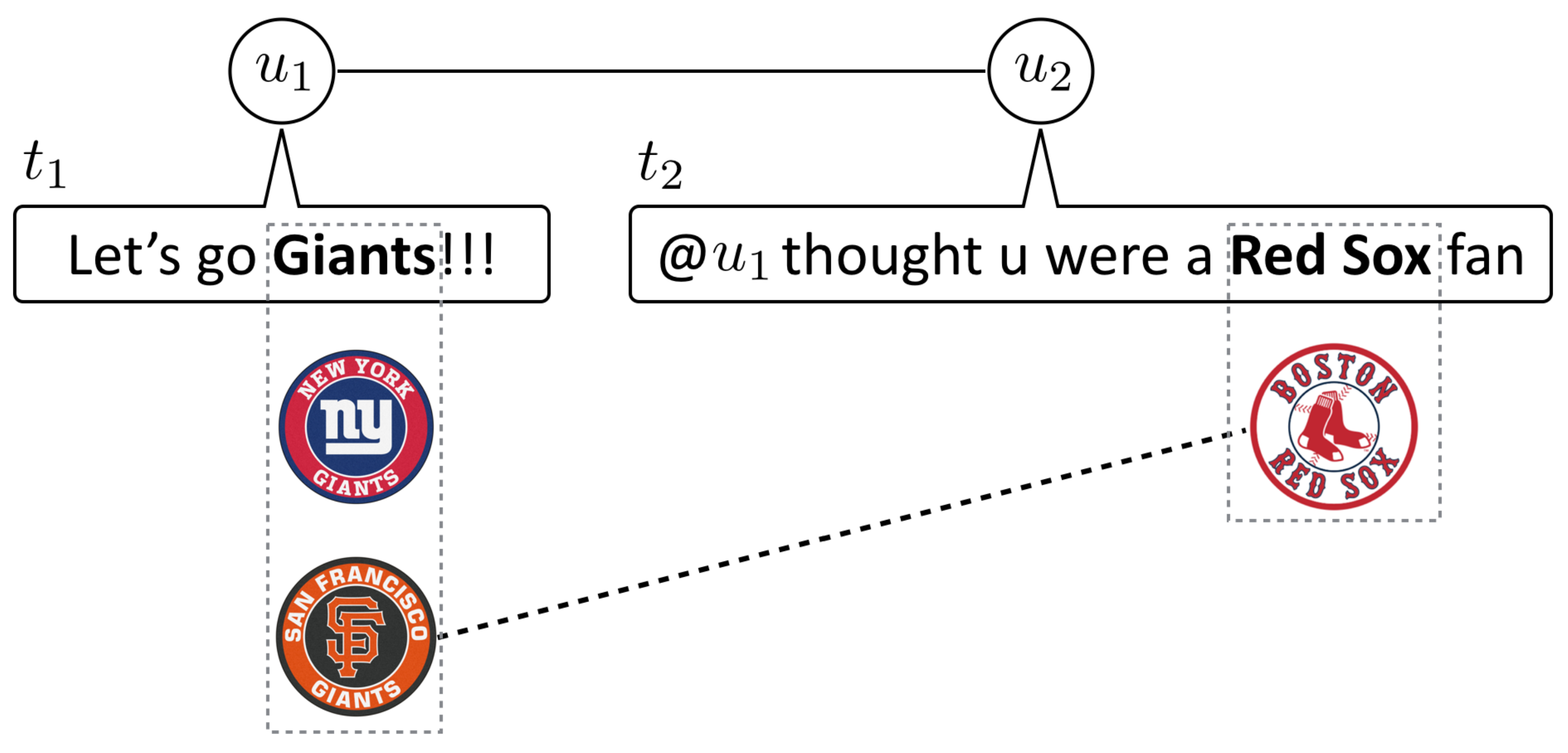}
\caption{Illustration on leveraging social relations for entity disambiguation. Socially connected users $u_1$ and $u_2$ tend to talk about similar entities (baseball in the example).}
\label{fig:example}
\end{figure}

Entity linking on short texts (e.g., Twitter messages) is of increasing interest, as it is an essential step for many downstream applications, such as market research~\cite{asur2010predicting}, topic detection and tracking~\cite{mathioudakis2010twittermonitor}, and question answering~\cite{yih2015semantic}. Tweet entity linking is a particularly difficult problem, because the short context around an entity mention is often insufficient for entity disambiguation. For example, as shown in Figure~\ref{fig:example}, the entity mention \example{Giants} in tweet $t_1$ can refer to the NFL football team \emph{New York Giants} or the MLB baseball team \emph{San Francisco Giants}. In this example, it is impossible to disambiguate between these entities solely based on the individual text message. 

We propose to overcome the difficulty and improve the entity disambiguation capability of the entity linking system by employing social network structures. The sociological theory of \emph{homophily} asserts that socially connected individuals are more likely to have similar behaviors or share similar interests~\cite{mcpherson2001birds}. This property has been used to improve many natural language processing tasks such as sentiment analysis~\cite{tan2011user,yang2015putting}, topic classification~\cite{hovy2015demographic} and user attribute inference~\cite{li2015learning}. We assume Twitter users will have similar interests in real world entities to their near neighbors --- an assumption of \emph{entity homophily} --- which is demonstrated in Figure~\ref{fig:example}. The social relation between users $u_1$ and $u_2$ may lead to more coherent topics in tweets $t_1$ and $t_2$. Therefore, by successfully linking the less ambiguous mention \example{Red Sox} in tweet $t_2$ to the \emph{Boston Red Sox} baseball team, the tweet entity linking system will be more confident on linking \example{Giants} to the \emph{San Francisco Giants} football team in tweet $t_1$. 

To exploit social information, we adopt the recent advance on embedding information networks~\cite{tang2015line}, which induces low-dimensional representations for author nodes based on the network structure. By learning the semantic interactions between the author embeddings and the pre-trained Freebase entity embeddings, the entity linking system can incorporate more disambiguating context from the social network. We also consider low-dimensional representations of mentions, another source of related information for entity linking, with the intuition that semantically related mentions can refer to similar entities. Previously proposed approaches~\cite{guo2013link,yang2015smart} are based on hand-crafted features and off-the-shelf machine learning algorithms. Our preliminary study suggests that simply augmenting the traditional surface features with the distributed representations barely improves the performance of these entity linking systems. Therefore, we propose \sysname, a {\bf N}eural model for {\bf T}weet {\bf E}ntity {\bf L}inking, to leverage the distributed representations of authors, mentions, and entities. \sysname\ can not only make efficient use of statistical surface features built from a knowledge base, but also learn the interactions between these distributed representations. 

Our contributions are summarized as follows:
\begin{itemize}
\item We present a novel model for entity linking that exploits distributed representations of users, mentions, and entities.
\item We combine this distributed model with a feedforward neural network that learns non-linear combinations of surface features.
\item We perform message-level inference using a dynamic program to avoid overlapping mentions. The architecture is trained with loss-augmented decoding, a large margin learning technique for structured prediction.
\item The complete system, \sysname, outperforms the previous state-of-the-art~\cite{yang2015smart} by 3\% average F1 on two benchmark datasets.
\end{itemize}

%% file: data.tex
\section{Data}
\label{sec:data}

\begin{table} [t]
\centering
\small\addtolength{\tabcolsep}{-2pt}
\begin{tabular}{lrrr}
    \toprule
    Data & \# Tweet & \# Entity & Date \\ \midrule
    NEEL-train & 2,340 & 2,202 & Jul. - Aug. 2011\\
    NEEL-test & 1,164 & 687 & Jul. - Aug. 2011 \\
    TACL & 500 & 300 & Dec. 2012 \\
    \bottomrule
\end{tabular}
\caption{Statistics of data sets.}
\label{tab:data}
\end{table}

Two publicly available datasets for tweet entity linking are adopted in the work. NEEL is originally collected and annotated for the Named Entity Extraction \& Linking Challenge~\cite{cano2014making}, and TACL is first used and released by~\newcite{fang2014entity}. The datasets are then cleaned and unified by~\newcite{yang2015smart}. The statistics of the datasets are presented in Table~\ref{tab:data}. 

%% file: homophily.tex
\section{Testing Entity Homophily}
\label{sec:homophily}

The hypothesis of \emph{entity homophily}, as presented in the introduction, is that socially connected individuals are more likely to mention similar entities than disconnected individuals. We now test the hypothesis on real data before we start building our entity linking systems. 

\paragraph{Twitter social networks}
We test the assumption on the users in the NEEL-train dataset. We construct three author social networks based on the follower, mention and retweet relations between the 1,317 authors in the NEEL-train dataset, which we refer as  \textsc{Follower}, \textsc{Mention} and  \textsc{Retweet}. Specifically, we use the Twitter API to crawl the friends of the NEEL users (individuals that they follow) and the mention/retweet links are induced from their most recent 3,200 tweets.\footnote{We are able to obtain at most 3,200 tweets for each Twitter user, due to the Twitter API limits.} We exploit bi-directed links to create the undirected networks, as bi-directed links result in stronger social network ties than directed links~\cite{kwak2010twitter,wu2011says}. The numbers of social relations for the networks are 1,604, 379 and 342 respectively.



\paragraph{Metrics}
We propose to use the \emph{entity-driven similarity} between authors to test the hypothesis of entity homophily. For a user $u_i$, we employ a Twitter NER system~\cite{ritter2011named} to detect entity mentions in the timeline, which we use to construct a user entity vector $\mb{u}_i^{(ent)}$, so that $u_{i,j}^{(ent)} = 1$ iff user $i$ has mentioned entity $j$.\footnote{We assume each name corresponds to a single entity for this metric, so this metric only approximates entity homophily.} The entity-driven similarity between two users $u_i$ and $u_j$ is defined as the cosine similarity score between the vectors $\mb{u}_i^{(ent)}$ and $\mb{u}_j^{(ent)}$. We evaluate the three networks by calculating the average entity-driven similarity of the connected user pairs and that of the disconnected user pairs, which we name as $sim(i \leftrightarrow j)$ and $sim(i \notleftright j)$.

\begin{table}
\centering
\small\addtolength{\tabcolsep}{-2pt}
\begin{tabular}{lrr}
    \toprule
    Network & $sim(i \leftrightarrow j)$ & $sim(i \notleftright j)$ \\ \midrule
    \textsc{Follower} & 0.128 & 0.025 \\
    \textsc{Mention} & 0.121 & 0.025 \\
    \textsc{Retweet} & 0.173 & 0.025 \\
    \bottomrule
\end{tabular}
\caption{The average entity-driven similarity results for the networks.}
\label{tab:network}
\end{table}

\paragraph{Results}
The entity-driven similarity results of these networks are presented in Table~\ref{tab:network}. As shown, $sim(i \leftrightarrow j)$ is substantially higher than $sim(i \notleftright j)$ on all three social networks, indicating that socially connected individuals clearly tend to mention more similar entities than disconnected individuals. Note that $sim(i \notleftright j)$ is approximately equal to the same base rate defined by the average entity-driven similarity of all pairs of users, because the vast majority of user pairs are disconnected, no matter how to define the network. Among the three networks, \textsc{Retweet} offers slightly higher $sim(i \leftrightarrow j)$ than \textsc{Follower} and \textsc{Mention}. The results verify our hypothesis of entity homophily, which forms the basis for this research.  Note that all social relation data was acquired in March 2016; by this time, the authorship information of 22.1\% of the tweets in the NEEL-train dataset was no longer available, because the tweets or user accounts had been deleted.

%% file: model.tex
\section{Method}
\label{sec:model}

\begin{figure}[t]
\centering
\includegraphics[scale=.3]{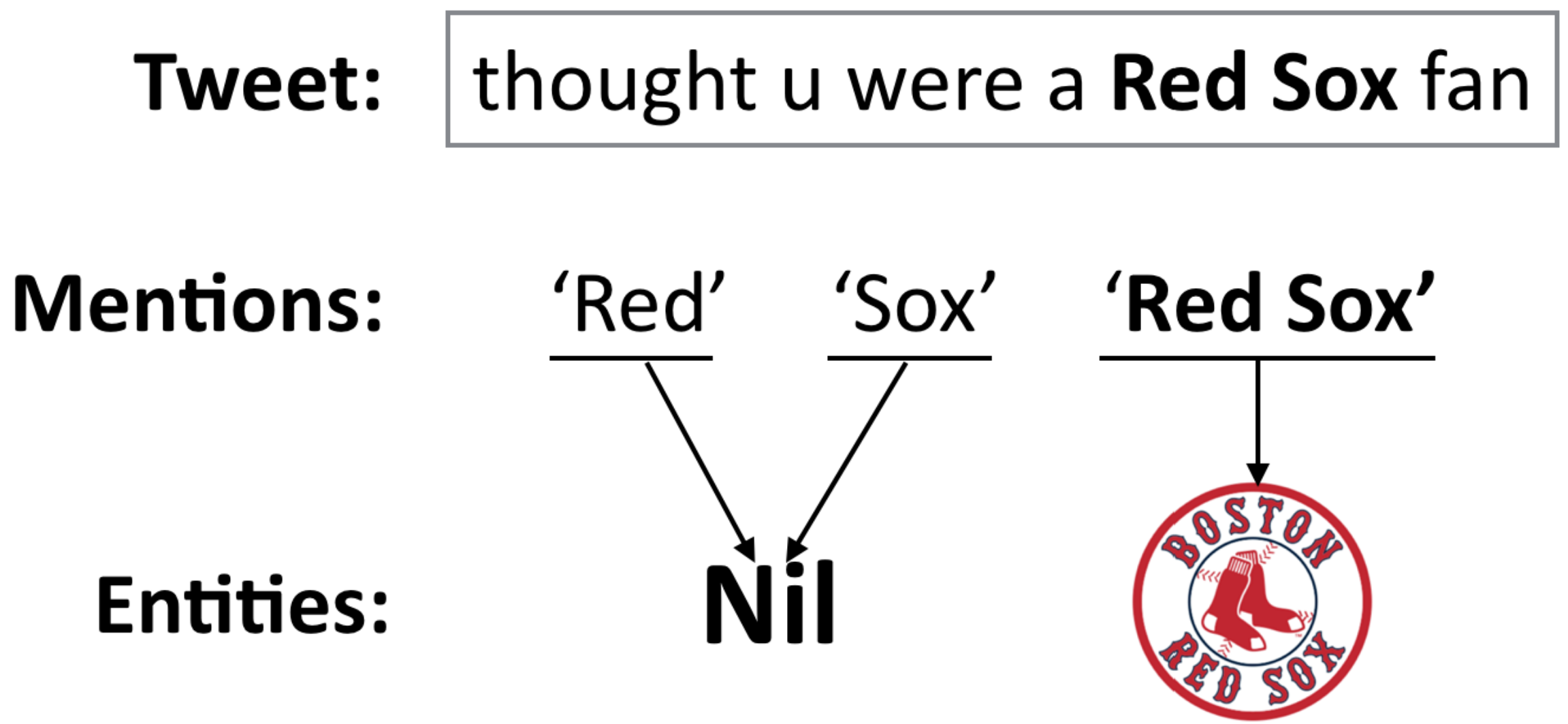}
\caption{Illustration of the non-overlapping structure for the task of tweet entity linking. In order to link \example{Red Sox} to a real entity, \example{Red} and \example{Sox} should be linked to {\bf Nil}.}
\label{fig:overlap}
\end{figure}

In this section, we present, \sysname, a novel neural based tweet entity linking framework that is able to leverage social information. We first formally define the task of tweet entity linking. Assume we are given an entity database (e.g., Wikipedia or Freebase), and a lexicon that maps a surface form into a set of entity candidates. For each input tweet, we consider any $n$-grams of the tweet that match the lexicon as mention candidates.\footnote{We adopted the same entity database and lexicon as those used by~\newcite{yang2015smart}. } The entity linking system maps every mention candidate (e.g., \example{Red Sox}) in the message to an entity (e.g., \emph{Boston Red Sox}) or to {\bf Nil} (i.e., not an entity). There are two main challenges in the problem. First, a mention candidate can often potentially link to multiple entities according to the lexicon. Second, as shown in~\autoref{fig:overlap}, many mention candidates overlap with each other. Therefore, the entity linking system is required to disambiguate entities and produce non-overlapping entity assignments with respect to the mention candidates in the tweet.

We formalize this task as a structured learning problem. Let $\mb{x}$ be the tweet, $u$ be the author, and $\mb{y}=\{y_t\}_{t=1}^T$ be the entity assignments of the $T$ mention candidates in the tweet. The overall scoring function $s(\mb{x}, \mb{y}, u)$ can be decomposed as follows,
\begin{equation}
s(\mb{x}, \mb{y}, u) = \sum_{t=1}^T g(\mb{x}, y_t, u, t),
\end{equation}
where $g(\mb{x}, y_t, u, t)$ is the scoring function for the $t$-th mention candidate choosing entity $y_t$.  Note that the system needs to produce non-overlapping entity assignments, which will be resolved in the inference algorithm. 

\begin{figure*}[t]
\centering
\includegraphics[scale=.48]{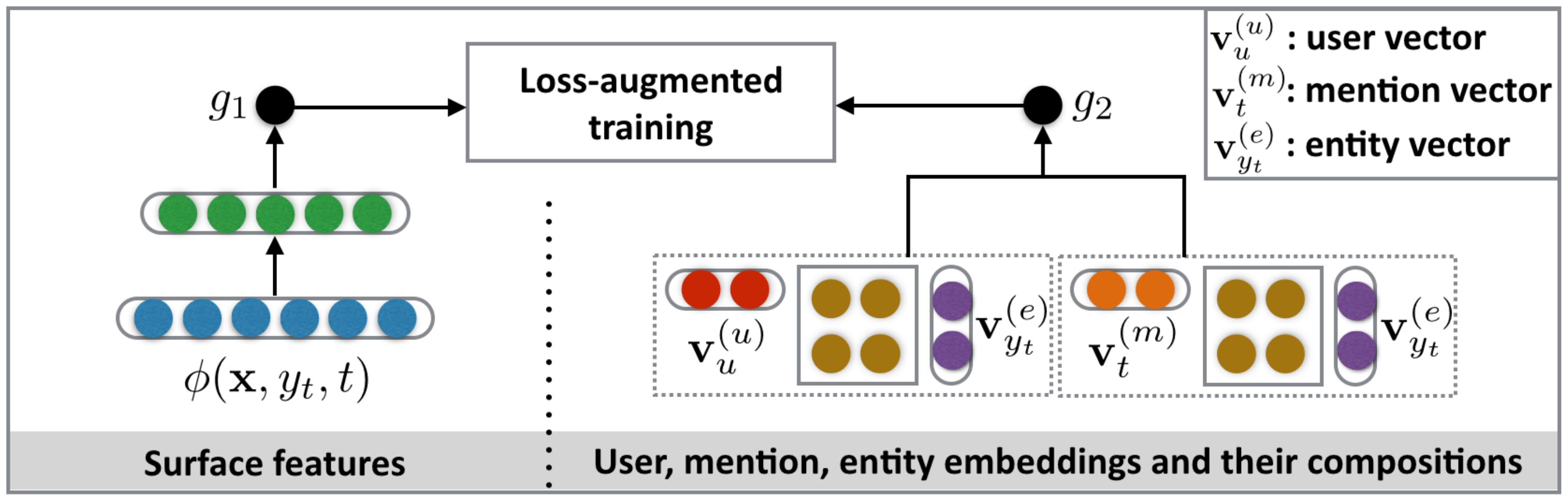}
\caption{The proposed neural network approach for tweet entity linking. A composition model based on bilinear functions is used to learn the semantic interactions of user, mention, and entity.}
\label{fig:overview}
\end{figure*}

The overview of \sysname~is illustrated in Figure~\ref{fig:overview}. We further break down $g(\mb{x}, y_t, u, t)$ into two scoring functions: 
\begin{align}
\notag g(\mb{x}, y_t, u, t; \Theta_1, \Theta_2) =& \\
 g_1 (\mb{x}, y_t, t; \Theta_1) &+ g_2 (\mb{x}, y_t, u, t; \Theta_2),
\end{align}
where $g_1$ is the scoring function for our basic surface features, and $g_2$ is the scoring function for modeling user, mention, entity representations and their compositions. $\Theta_1$ and $\Theta_2$ are model parameters that will be detailed below.
We choose to use a multilayer perceptron (MLP) to model $g_1 (\mb{x}, y_t, t; \Theta_1)$, and we employ simple yet efficient bilinear functions to learn the compositions of user, mention, and entity representations $g_2 (\mb{x}, y_t, u, t; \Theta_2)$. Finally, we present a training algorithm based on loss-augmented decoding and a non-overlapping inference algorithm.

\subsection{Modeling Surface Features}
We include the 37 features used by~\newcite{yang2015smart} as our surface feature set. These features are extracted from various sources, including a named entity recognizer, an entity type recognizer, and some statistics of the Wikipedia pages.


We exploit a multilayer perceptron (MLP) to transform the surface features to a real-valued score. The output of the MLP is formalized as follows,
\begin{align}
\notag g_1 (\mb{x}, y_t, t; \Theta_1) =& \bm{\beta}^\top \mb{h} + b \\
\mb{h} =& \text{tanh}(\mb{W} \phi (\mb{x}, y_t, t) + \mb{b}),
\end{align}
where $\phi (\mb{x}, y_t, t)$ is the feature function, $\mb{W}$ is an $M \times D$ matrix, the weights $\mb{b}$ are bias terms, and $\mb{h}$ is the output of the hidden layer of the MLP. $\bm{\beta}$ is an $M$ dimensional vector of weights for the output score, and $b$ is the bias term. The parameters of the MLP are $\Theta_1 = \{ \mb{W}, \mb{b},  \bm{\beta}, b \}$.
\newcite{yang2015smart} argue that non-linearity is the key for obtaining good results on the task, as linear models are not expressive enough to capture the high-order relationships between the dense features. They propose a tree-based non-linear model for the task. The MLP forms simple non-linear mappings between the input features and the output score, whose parameters will be jointly learnt with other components in \sysname.

\subsection{Modeling User, Mention, and Entity}
To leverage the social network structure, we first train low-dimensional embeddings for the authors using the social relations. The mention and entity representations are given by word embeddings learnt with a large Twitter corpus and pre-trained Freebase entity embeddings respectively. We will denote the user, word, entity embedding matrices as:
\[ \mb{E}^{(u)} = \{ \mb{v}_u^{(u)} \} \text{ \hspace{0.4cm}  } \mb{E}^{(w)} = \{ \mb{v}_w^{(w)} \} \text{ \hspace{0.4cm} } \mb{E}^{(e)} = \{ \mb{v}_e^{(e)} \}, \]
where $\mb{E}^{(u)}, \mb{E}^{(w)}, \mb{E}^{(e)}$ are $V^{(u)} \times D^{(u)}, V^{(w)} \times D^{(w)}, V^{(e)} \times D^{(e)}$ matrices, and $\mb{v}_u^{(u)}$, $\mb{v}_w^{(w)}$, $\mb{v}_e^{(e)}$ are $D^{(u)}$, $D^{(w)}$, $D^{(e)}$ dimensional embedding vectors respectively. $V^{(u)}, V^{(w)}, V^{(e)}$ are the vocabulary sizes for users, words, and entities.
Finally, we present a composition model for learning semantic interactions between user, mention, and entity.

\paragraph{User embeddings}
We obtain low-dimensional Twitter author embeddings $\mb{E}^{(u)}$ using \emph{LINE} --- the recently proposed model for embedding information networks~\cite{tang2015line}. Specifically, we train \emph{LINE} with the second-order proximity, which assumes that Twitter users sharing many neighbors are close to each other in the embedding space. According to the original paper, the second-order proximity yields slightly better performances than the first-order proximity, which assumes connecting users are close to each other, on a variety of downstream tasks.

\paragraph{Mention embeddings}
The representation of a mention is the average of embeddings of words it contains. As each mention is typically one to three words, the simple representations often perform surprisingly well~\cite{SocherChenManningNg2013}. We adopt the structured skip-gram model~\cite{ling2015two} to learn the word embeddings $\mb{E}^{(w)}$ on a Twitter corpus with 52 million tweets~\cite{owoputi2013improved}. The mention vector of the $t$-th mention candidate can be written as:
\begin{equation}
\mb{v}_t^{(m)} = \frac{1}{|\mb{x}_t^{(w)}|} \sum_{w \in \mb{x}_t^{(w)}} \mb{v}_w^{(w)},
\end{equation}
where $\mb{x}_t^{(w)}$ is the set of words in the mention.

\paragraph{Entity embeddings}
We use the pre-trained Freebase entity embeddings released by Google to represent entity candidates, which we refer as $\mb{E}^{(e)}$.\footnote{Available at https://code.google.com/archive/p/word2vec/} The embeddings are trained with the skip-gram model~\cite{mikolov2013distributed} on 100 billion words from various news articles. The entity embeddings can also be learnt from Wikipedia hyperlinks or Freebase entity relations, which we leave as future work.

\paragraph{Compositions of user, mention, and entity}
The distributed representations of users, mentions, and entities offer additional information that is useful for improving entity disambiguation capability. In particular, we explore the information by making two assumptions: socially connected users are interested in similar entities (entity homophily), and semantically related mentions are likely to be linked to similar entities. 

We utilize a simple composition model that takes the form of the summation of two bilinear scoring functions, each of which explicitly leverages one of the assumptions. Given the author representation $\mb{v}_u^{(u)} $, the mention representation $\mb{v}_t^{(m)}$, and the entity representation $\mb{v}_{y_t}^{(e)}$, the output of the model can be written as:
\begin{align}
\notag g_2 (\mb{x}, y_t, u, t; \Theta_2) =& {\mb{v}_u^{(u)}}^\top \mb{W}^{(u,e)} \mb{v}_{y_t}^{(e)} \\
&+ {\mb{v}_t^{(m)}}^\top \mb{W}^{(m,e)} \mb{v}_{y_t}^{(e)}, \label{eq:bilinear}
\end{align}
where $\mb{W}^{(u,e)}$ and $\mb{W}^{(m,e)}$ are $D^{(u)} \times D^{(e)}$ and $D^{(w)} \times D^{(e)}$ bilinear transformation matrices. Similar bilinear formulation has been used in the literature of knowledge base completion and inference~\cite{SocherChenManningNg2013,yang2014embedding}. The parameters of the composition model are $\Theta_2 = \{ \mb{W}^{(u,e)}, \mb{W}^{(m,e)}, \mb{E}^{(u)}, \mb{E}^{(w)}, \mb{E}^{(e)} \}$.

\newcommand{\nil}[0]{\textbf{Nil}}
\subsection{Non-overlapping Inference}
The non-overlapping constraint for entity assignments requires inference method that is different from the standard Viterbi algorithm for a linear chain. We now present a variant of the Viterbi algorithm for the non-overlapping structure. Given the overall scoring function $g(\mb{x}, y_t, u, t)$ for the $t$-th mention candidate choosing an entity $y_t$, we sort the mention candidates by their end indices and define the Viterbi recursion by 
\begin{align}
\hat{y_t} =& \argmax_{  y_t \in \mathcal{Y}_{\mb{x}_t}, y_t \neq  \nil} g(\mb{x}, y_t, u, t)\\
 a(1) =& \max(g(\mb{x}, \nil, u, 1), g(\mb{x}, \hat{y_1}, u, 1)) \\
 a(t) =& \max \left( \psi_t(\nil), \psi_t(\hat{y}_t) \right)\\
\psi_t(\nil) = & g(\mb{x},\nil,u,t) + a(t-1)\\
\notag
\psi_t(\hat{y}_t) = & g(\mb{x},\hat{y}_t,u,t) + \sum_{prev(t)<t'< t} g(\mb{x}, \nil, u, t')\\
& + a(prev(t))
\end{align}
where $\mathcal{Y}_{\mb{x}_t}$ is set of entity candidates for the $t$-th mention candidate, and
$prev(t)$ is a function that points out the previous non-overlapping mention candidate for the $t$-th mention candidate.
We exclude any second-order features between entities. Therefore, for each mention candidate, we only need to decide whether it can take the highest scored entity candidate $\hat{y_t} $ or the special {\bf Nil} entity based on whether it is overlapped with other mention candidates. 

\subsection{Loss-augmented Training}
The parameters need to be learnt during training are $\Theta = [\Theta_1, \{ \mb{W}^{(u,e)}, \mb{W}^{(m,e)} \}]$.\footnote{We fixed the pre-trained embedding matrices during loss-augmented training.} We train \sysname~by minimizing the following loss function for each training tweet:
\begin{equation}
L (\Theta) = \max_{\mb{y} \in \mathcal{Y}_\mb{x}} (\Delta (\mb{y}, \mb{y}^*) + s(\mb{x}, \mb{y}, u) ) - s(\mb{x}, \mb{y}^*, u),
\label{eq:loss}
\end{equation}
where $\mb{y}^*$ is the gold structure, $\mathcal{Y}_\mb{x}$ represents the set of valid output structures for $\mb{x}$, and $\Delta (\mb{y}, \mb{y}^*)$ is the weighted hamming distance between the gold structure $\mb{y}^*$ and the valid structure $\mb{y}$. 
The hamming loss is decomposable on the mention candidates, which enables efficient inferences. 
We set the hamming loss weight to $0.2$ after a preliminary search. Note that the number of parameters in our composition model is large. Thus, we include an L2 regularizer on these parameters, which is omitted from~\autoref{eq:loss} for brevity. The evaluation of the loss function corresponds to the loss-augmented inference problem:
\begin{equation}
\hat{\mb{y}} = \argmax_{\mb{y} \in \mathcal{Y}_\mb{x}} (\Delta (\mb{y}, \mb{y}^*) + s(\mb{x}, \mb{y}, u) ),
\end{equation}
which can be solved by the above non-overlapping inference algorithm.
We employ vanilla SGD algorithm to optimize all the parameters. The numbers of training epochs are determined by early stopping (at most 1000 epochs). Training takes 6-8 hours on 4 threads.

%% file: exp.tex
\section{Experiments}
\label{sec:exp}

In this section, we evaluate \sysname~on the NEEL and TACL datasets as described in~\autoref{sec:data}, focusing on investigating whether social information can improve the task. We also compare \sysname~with the previous state-of-the-art system.

\subsection{Social network expansion}
We utilize Twitter follower, mention, and retweet social networks to train user embeddings. We were able to identify 2,312 authors for the tweets of the two datasets in March 2016. We then used the Twitter API to crawl their friend links and timelines, from which we can induce the networks.  We find the numbers of social connections (bidirectional links) between these users are relatively small. In order to learn better user embeddings, we expand the set of author nodes by including nodes that will do the most to densify the author networks. For the follower network, we add additional individuals who are followed by at least twenty authors in the original set. For the mention or retweet networks, we add all users who have mentioned or retweeted by at least ten authors in the original set. The statistics of the resulting networks are presented in Table~\ref{tab:networks}.

\begin{table}
\centering
\small\addtolength{\tabcolsep}{-2pt}
\begin{tabular}{lrr}
    \toprule
    Network & \# Author & \# Relation  \\ \midrule
    \textsc{Follower+} & 8,772 & 286,800  \\
    \textsc{Mention+} & 6,119 & 57,045 \\
    \textsc{Retweet+} & 7,404 & 59,313 \\
    \bottomrule
\end{tabular}
\caption{Statistics of author social networks used for training user embeddings.}
\label{tab:networks}
\end{table}

\subsection{Experimental Settings}
Following \newcite{yang2015smart}, we train all the models with the NEEL-train dataset and evaluate different systems on the NEEL-test and TACL datasets. In addition, 800 tweets from the NEEL-train dataset are sampled as our development set to perform parameter tuning. Note that~\newcite{yang2015smart} also attempt to optimize F1 scores by balancing precision and recall scores on the development set; we do not fine tune our F1 in this way, so that we can apply a single trained system across different test sets.


\paragraph{Metrics}
We follow prior work~\cite{guo2013link,yang2015smart} and perform the standard evaluation for an end-to-end entity linking system, computing precision, recall, and F1 score according to the entity references and the system outputs. An output entity is considered as correct if it matches the gold entity and the mention boundary overlaps with the gold mention boundary. More details about the metrics are described by~\newcite{carmel2014erd}.

\paragraph{Competitive systems}
Our first baseline system, \sysname-nonstruct, ignores the structure information and makes the entity assignment decision for each mention candidate individually. For \sysname, we start with a baseline system using the surface features, and then incorporate the two bilinear functions (user-entity and mention-entity) described in~\autoref{eq:bilinear} incrementally. Our main evaluation uses the \textsc{Retweet+} network, since the retweet network had the greatest entity homophily; an additional evaluation compares across network types.

\paragraph{Parameter tuning}
We tune all the hyper-parameters on the development set, and then re-train the models on the full training data with the best parameters. We choose the number of hidden units for the MLP from $\{ 20, 30, 40, 50 \}$, and the regularization penalty for our composition model from $\{ 0.001, 0.005, 0.01, 0.05, 0.1 \}$. The sizes of user embeddings and word embeddings are selected from $\{ 50, 100 \}$ and $\{ 200, 400, 600 \}$ respectively. The pre-trained Freebase entity embedding size is $1000$. The learning rate for the SGD algorithm is set as $0.01$. During training, we check the performance on the development set regularly to perform early stopping.

\subsection{Results}

\input{updated-results-tab}

Table~\ref{tab:results} summarizes the empirical findings for our approach and S-MART~\cite{yang2015smart} on the tweet entity linking task.  For the systems with user-entity bilinear function, we report results obtained from embeddings trained on \textsc{Retweet+} in~\autoref{tab:results}, and other results are available in~\autoref{tab:user}. The best hyper-parameters are: the number of hidden units for the MLP is $40$,  the L2 regularization penalty for the composition parameters is $0.005$, and the user embedding size is $100$. For the word embedding size, we find $600$ offers marginal improvements over $400$ but requires longer training time. Thus, we choose $400$ as the size of word embeddings.

As presented in~\autoref{tab:results}, \sysname-nonstruct performs 2.7\% F1 worse than the \sysname~baseline on the two test sets, which indicates the non-overlapping inference improves system performance on the task. With structured inference but without embeddings, \sysname\ performs roughly the same as S-MART, showing that a feedforward neural network offers similar expressivity to the regression trees employed by \newcite{yang2015smart}.


Performance improves substantially with the incorporation of low-dimensional author, mention, and entity representations. As shown in Table~\ref{tab:results}, by learning the interactions between mention and entity representations, \sysname~with mention-entity bilinear function outperforms the  \sysname~baseline system by 1.8\% F1 on average. Specifically, the bilinear function results in considerable performance gains in recalls, with small compromise in precisions on the datasets.

Social information helps to increase about 1\% F1 on top of both the \sysname~baseline system and the \sysname~system with mention-entity bilinear composition. In contrast to the mention-entity composition model, which mainly focuses on improving the baseline system on recall scores, the user-entity composition model increases around 2.5\% recalls, without much sacrifice in precisions.

Our best system achieves the state-of-the-art results on the NEEL-test dataset and the TACL dataset, outperforming S-MART by 0.9\% and 5.4\% F1 scores respectively. To establish the statistical significance of the results, we obtain 100 bootstrap samples for each test set, and compute the F1 score on each sample for each algorithm.  Two-tail paired t-test is then applied to determine if the F1 scores of two algorithms are significantly different. \sysname\ significantly outperforms S-MART on the NEEL-test dataset and the TACL dataset under $p < 0.01$ level, with t-statistics equal to $11.5$ and $33.6$ respectively.

As shown in~\autoref{tab:user}, \textsc{Mention+} and \textsc{Retweet+} perform slightly better than \textsc{Follower+}. \newcite{puniyani2010social} show that the mention network has stronger linguistic properties than the follower network, as it gives better correlations on each author's distribution over latent topics as induced by latent Dirichlet allocation~\cite{blei2003latent}. Our results suggest that the properties hold with respect to the authors' interests on real world entities. 

\subsection{Error Analysis \& Discussion}
We examine the outputs of different systems, focusing on investigating what errors are corrected by the two bilinear functions. The results reveal that the mention-entity composition improves the system ability to tackle mentions that are abbreviations such as  \example{WSJ} (\emph{The Wall Street Journal}) and \example{SJSU} (\emph{San Jose State University}), which leads to higher recall scores. The mention-entity model also helps to eliminate errors that incorrectly link non-entities to popular entities. For example, the \sysname~baseline system links \example{sec} in the tweet \example{I'm a be in Miami for sec to hit da radio!}  to \emph{Southeastern Conference}, which is corrected by the mention-entity composition model. The word semantic information encoded in the mention representations alleviates the biased entity information given by the surface features.

The user-entity composition model is good at handling highly ambiguous mentions. For example, our full model successfully disambiguates entities for mentions such as \example{Sox} (\emph{Boston Red Sox} vs.  \emph{Chicago White Sox}),  \example{Sanders} (\emph{Bernie Sanders} vs.  \emph{Barry Sanders}), and \example{Memphis} (\emph{Memphis Grizzlies} vs.  \emph{Memphis, Tennessee}),  which are mistakenly linked to the other entities or {\bf Nil} by the mention-entity model. Another example is that the social network information helps the system correctly link \example{Kim} to \emph{Lil' Kim} instead of \emph{Kim Kardashian}, despite that the latter entity's wikipedia page is considerably more popular.

%% file: updated-results-tab.tex
\begin{table*} [ht!]
\centering
\small
\begin{tabular}{llllllllll}
    \toprule
    \multirow{2}{*}{System} & \multirow{2}{*}{\parbox{1.2cm}{user\\-entity}} & \multirow{2}{*}{\parbox{1.2cm}{mention\\-entity}} & \multicolumn{3}{c}{NEEL-test} & \multicolumn{3}{c}{TACL} & \multirow{2}{*}{Avg. F1}\\
    \cmidrule(l){4-6} \cmidrule(l){7-9}
          & & & P & R & F1 & P & R & F1 &  \\ \midrule
   \multicolumn{10}{l}{\it Our approach} \\
   \sysname-nonstruct & & & 80.0 & 68.0 & 73.5 & 64.7 & 62.3 & 63.5 & 68.5 \\[3pt]
   \sysname          & & & {\bf 82.8} & 69.3 & 75.4 & 68.0 & 66.0 & 67.0 & 71.2 \\
 \sysname & \checkmark & & 82.3 & 71.8 & 76.7 & 66.9 & 68.7 & 67.8 & 72.2 \\
   \sysname & & \checkmark & 80.2 & {\bf 75.8} & 77.9 & 66.9 & {\bf 69.3} & 68.1 & 73.0 \\
   \sysname & \checkmark & \checkmark & 81.9 & 75.6 & {\bf 78.6} & {\bf 69.0} & 69.0 & {\bf 69.0}  & {\bf 73.8} \\[6pt]
   \multicolumn{10}{l}{\it Best published results} \\
     S-MART & & & 80.2 & 75.4 & 77.7 & 60.1 & 67.7 & 63.6 & 70.7  \\
    \bottomrule
\end{tabular}
\caption{Evaluation results on the NEEL-test and TACL datasets for different systems. The best results are in {\bf bold}.}
\label{tab:results}
\end{table*}

\begin{table} [ht!]
\centering
\small
\addtolength{\tabcolsep}{-2pt}
\begin{tabular}{lllllll}
    \toprule
    \multirow{2}{*}{Network} & \multicolumn{3}{c}{NEEL-test} & \multicolumn{3}{c}{TACL}\\
    \cmidrule(l){2-4} \cmidrule(l){5-7}
           & P & R & F1 & P & R & F1  \\ \midrule
     \textsc{Follower+} & 82.2 & 75.1 & 78.5 & 67.8 & 68.7 & 68.2  \\
    \textsc{Mention+}  & {\bf 82.5} & {\bf 76.0} & {\bf 79.1} & 67.5 & {\bf 69.3} & 68.4 \\
    \textsc{Retweet+}  & 81.9 & 75.6 & 78.6 & {\bf 69.0} & 69.0 & {\bf 69.0}  \\
    \bottomrule
\end{tabular}
\caption{Comparison of different social networks with our full model. The best results are in {\bf bold}.}
\label{tab:user}
\end{table}

%% file: related.tex
\section{Related Work}
\label{sec:related}

\paragraph{Tweet entity linking}
Previous work on entity linking mainly focuses on well-written documents~\cite{bunescu2006using,cucerzan2007large,milne2008learning}, where entity disambiguation is usually performed by maximizing the global topical coherence between entities. However, these approaches often yield unsatisfactory performance on Twitter messages, due to the short and noisy nature of the tweets. To tackle this problem, collective tweet entity linking methods that leverage enriched context and metadata information have been proposed~\cite{huang2014collective}. \newcite{guo2013microblog} search for textually similar tweets for a target tweet, and encourage these Twitter messages to contain similar entities through label propagation. \newcite{shen2013linking} employ Twitter user account information to improve entity linking, based on the intuition that all tweets posted by the same user share an underlying topic distribution. \newcite{fang2014entity} demonstrate that spatial and temporal signals are critical for the task, and they advance the performance by associating entity prior distributions with different timestamps and locations. Our work overcomes the difficulty by leveraging social relations --- socially connected individuals are assumed to share similar interests on entities. As the Twitter post information is often sparse for some users, our assumption enables the utilization of more relevant information that helps to improve the task.

\paragraph{NLP with social relations}
Most previous work on incorporating social relations for NLP problems focuses on Twitter sentiment analysis, where the existence of social relations between users is considered as a clue that the sentiment polarities of messages from the users should be similar. \newcite{speriosu2011twitter} construct a heterogeneous network with tweets, users, and n-grams as nodes, and the sentiment label distributions associated with the nodes are refined by performing label propagation over social relations. \newcite{tan2011user} and~\newcite{hu2013exploiting} leverage social relations for sentiment analysis by exploiting a factor graph model and the graph Laplacian technique respectively, so that the tweets belonging to social connected users share similar label distributions. We work on entity linking in Twitter messages, where the label space is much larger than that of sentiment classification. The social relations can be more relevant in our problem, as it is challenging to obtain the entity prior distribution for each individual.

%% file: con.tex
\section{Conclusion}
We present a neural based structured learning architecture for tweet entity linking, leveraging the tendency of socially linked individuals to share similar interests on named entities --- the phenomenon of \emph{entity homophily}. By modeling the compositions of vector representations of author, entity, and mention, our approach is able to exploit the social network as a source of contextual information. This vector-compositional model is combined with non-linear feature combinations of surface features, via a feedforward neural network. To avoid predicting overlapping entity mentions, we employ a structured prediction algorithm, and train the system with loss-augmented decoding. 

Social networks arise in other settings besides microblogs, such as webpages and academic research articles; exploiting these networks is a possible direction for future work. We would also like to investigate other metadata attributes that are relevant to the task, such as spatial and temporal signals. 

%% file: ack.tex
\paragraph{Acknowledgments}
This research was supported by the National Science Foundation under awards IIS-1111142 and RI-1452443, by the National Institutes of Health under award number R01-GM112697-01, and by the Air Force Office of Scientific Research.

%% file: appendix.tex
\onecolumn
\begin{appendices}
\section{Appendix: Additional Results}
\label{app:results}

\input{results-tab}

In the first version of \cite{yang2015smart}, the Twitter messages that contain no ground truth entities are excluded in the experiments. For completeness, we now present the evaluation results of \sysname~in this setting, which are shown in Table~\ref{tab:app-results}. The \textsc{Retweet+} network is adopted to train author embeddings. The best hyper-parameters are the same as those described in~\autoref{sec:exp}, except for the L2 regularization penalty for the composition parameters, which is set as $0.01$ here.

The results are generally better than those presented in Table~\ref{tab:results}. As shown, \sysname~benefits from the distributed representations of authors, mentions, and entities, which improve the average F1 score by 2.3 points. \sysname~also gives the best results on the datasets, outperforming S-MART by about 2\% F1 on average.

\end{appendices}

%% file: results-tab.tex
\begin{table*} [ht!]
\centering
\small
\begin{tabular}{llllllllll}
    \toprule
    \multirow{2}{*}{System} & \multirow{2}{*}{\parbox{1.2cm}{user\\-entity}} & \multirow{2}{*}{\parbox{1.2cm}{mention\\-entity}} & \multicolumn{3}{c}{NEEL-test} & \multicolumn{3}{c}{TACL} & \multirow{2}{*}{Avg. F1}\\
    \cmidrule(l){4-6} \cmidrule(l){7-9}
          & & & P & R & F1 & P & R & F1 &  \\ \midrule
   \multicolumn{10}{l}{\it Our approach} \\
     \sysname-nonstruct & & & 83.0 & 71.8 & 77.0 & 80.9 & 69.0 & 74.5 & 75.8 \\[3pt]
   \sysname & & & 84.4 & 73.9 & 78.8 & 82.0 & 71.3 & 76.3 & 77.6\\
 \sysname & \checkmark & & 83.8 & 76.7 & 80.1 & 81.8 & {\bf 73.3} & 77.3 & 78.7 \\
  \sysname   &  & \checkmark & 84.1 & 78.3 & 81.1 & 83.0 & 71.7 & 76.9 & 79.0 \\
\sysname     & \checkmark & \checkmark &{\bf 84.8} & {\bf 79.3} & {\bf 82.0} & {\bf 83.5} & 72.7 & {\bf 77.7}  & {\bf 79.9} \\[6pt]
   \multicolumn{10}{l}{\it Best published results} \\
     S-MART & & & 83.2 & 79.2 & 81.1 & 76.8 & 73.0 & 74.9 & 78.0  \\
    \bottomrule
\end{tabular}
\caption{Evaluation results on the NEEL-test and TACL datasets for different systems. Twitter messages that contain no ground truth entities are excluded for both training and testing. The best results are in {\bf bold}.}
\label{tab:app-results}
\end{table*}

%% file: main.bbl
\begin{thebibliography}{}

\bibitem[\protect\citename{Asur and Huberman}2010]{asur2010predicting}
Sitaram Asur and Bernardo~A Huberman.
\newblock 2010.
\newblock Predicting the future with social media.
\newblock In {\em Web Intelligence and Intelligent Agent Technology (WI-IAT)},
  pages 492--499.

\bibitem[\protect\citename{Blei \bgroup et al.\egroup }2003]{blei2003latent}
David~M Blei, Andrew~Y Ng, and Michael~I Jordan.
\newblock 2003.
\newblock Latent dirichlet allocation.
\newblock {\em the Journal of machine Learning research}, 3:993--1022.

\bibitem[\protect\citename{Bunescu and Pasca}2006]{bunescu2006using}
R.~C Bunescu and M.~Pasca.
\newblock 2006.
\newblock Using encyclopedic knowledge for named entity disambiguation.
\newblock In {\em {Proceedings of the European Chapter of the Association for
  Computational Linguistics (EACL)}}.

\bibitem[\protect\citename{Cano \bgroup et al.\egroup }2014]{cano2014making}
Amparo~E Cano, Giuseppe Rizzo, Andrea Varga, Matthew Rowe, Milan Stankovic, and
  Aba-Sah Dadzie.
\newblock 2014.
\newblock Making sense of microposts (\# microposts2014) named entity
  extraction \& linking challenge.
\newblock {\em Making Sense of Microposts (\# Microposts2014)}.

\bibitem[\protect\citename{Carmel \bgroup et al.\egroup }2014]{carmel2014erd}
David Carmel, Ming-Wei Chang, Evgeniy Gabrilovich, Bo-June~Paul Hsu, and
  Kuansan Wang.
\newblock 2014.
\newblock Erd'14: entity recognition and disambiguation challenge.
\newblock In {\em ACM SIGIR Forum}, pages 63--77.

\bibitem[\protect\citename{Cucerzan}2007]{cucerzan2007large}
Silviu Cucerzan.
\newblock 2007.
\newblock Large-scale named entity disambiguation based on wikipedia data.
\newblock In {\em {Proceedings of Empirical Methods for Natural Language
  Processing (EMNLP)}}.

\bibitem[\protect\citename{Fang and Chang}2014]{fang2014entity}
Yuan Fang and Ming-Wei Chang.
\newblock 2014.
\newblock Entity linking on microblogs with spatial and temporal signals.
\newblock {\em Transactions of the Association for Computational Linguistics
  (ACL)}.

\bibitem[\protect\citename{Guo \bgroup et al.\egroup }2013a]{guo2013link}
Stephen Guo, Ming-Wei Chang, and Emre Kiciman.
\newblock 2013a.
\newblock To link or not to link? a study on end-to-end tweet entity linking.
\newblock In {\em {Proceedings of the North American Chapter of the Association
  for Computational Linguistics (NAACL)}}, Atlanta, GA.

\bibitem[\protect\citename{Guo \bgroup et al.\egroup }2013b]{guo2013microblog}
Yuhang Guo, Bing Qin, Ting Liu, and Sheng Li.
\newblock 2013b.
\newblock Microblog entity linking by leveraging extra posts.
\newblock In {\em {Proceedings of Empirical Methods for Natural Language
  Processing (EMNLP)}}, Seattle, WA.

\bibitem[\protect\citename{Hovy}2015]{hovy2015demographic}
Dirk Hovy.
\newblock 2015.
\newblock Demographic factors improve classification performance.
\newblock In {\em {Proceedings of the Association for Computational Linguistics
  (ACL)}}, pages 752--762, Beijing, China.

\bibitem[\protect\citename{Hu \bgroup et al.\egroup }2013]{hu2013exploiting}
Xia Hu, Lei Tang, Jiliang Tang, and Huan Liu.
\newblock 2013.
\newblock Exploiting social relations for sentiment analysis in microblogging.
\newblock In {\em Proceedings of the sixth ACM international conference on Web
  search and data mining {(WSDM)}}, pages 537--546.

\bibitem[\protect\citename{Huang \bgroup et al.\egroup
  }2014]{huang2014collective}
Hongzhao Huang, Yunbo Cao, Xiaojiang Huang, Heng Ji, and Chin-Yew Lin.
\newblock 2014.
\newblock Collective tweet wikification based on semi-supervised graph
  regularization.
\newblock In {\em {Proceedings of the Association for Computational Linguistics
  (ACL)}}, Baltimore, MD.

\bibitem[\protect\citename{Kwak \bgroup et al.\egroup }2010]{kwak2010twitter}
Haewoon Kwak, Changhyun Lee, Hosung Park, and Sue Moon.
\newblock 2010.
\newblock What is {Twitter}, a social network or a news media?
\newblock In {\em {Proceedings of the Conference on World-Wide Web (WWW)}},
  pages 591--600, New York. {ACM}.

\bibitem[\protect\citename{Li \bgroup et al.\egroup }2015]{li2015learning}
Jiwei Li, Alan Ritter, and Dan Jurafsky.
\newblock 2015.
\newblock Learning multi-faceted representations of individuals from
  heterogeneous evidence using neural networks.
\newblock {\em arXiv preprint arXiv:1510.05198}.

\bibitem[\protect\citename{Ling \bgroup et al.\egroup }2015]{ling2015two}
Wang Ling, Chris Dyer, Alan Black, and Isabel Trancoso.
\newblock 2015.
\newblock Two/too simple adaptations of word2vec for syntax problems.
\newblock In {\em {Proceedings of the North American Chapter of the Association
  for Computational Linguistics (NAACL)}}, Denver, CO.

\bibitem[\protect\citename{Mathioudakis and
  Koudas}2010]{mathioudakis2010twittermonitor}
Michael Mathioudakis and Nick Koudas.
\newblock 2010.
\newblock Twittermonitor: trend detection over the twitter stream.
\newblock In {\em Proceedings of the ACM SIGMOD International Conference on
  Management of data (SIGMOD)}, pages 1155--1158.

\bibitem[\protect\citename{McPherson \bgroup et al.\egroup
  }2001]{mcpherson2001birds}
Miller McPherson, Lynn Smith-Lovin, and James~M Cook.
\newblock 2001.
\newblock Birds of a feather: Homophily in social networks.
\newblock {\em Annual review of sociology}, pages 415--444.

\bibitem[\protect\citename{Mikolov \bgroup et al.\egroup
  }2013]{mikolov2013distributed}
Tomas Mikolov, Ilya Sutskever, Kai Chen, Greg~S Corrado, and Jeff Dean.
\newblock 2013.
\newblock Distributed representations of words and phrases and their
  compositionality.
\newblock In {\em {Neural Information Processing Systems (NIPS)}}, pages
  3111--3119, Lake Tahoe.

\bibitem[\protect\citename{Milne and Witten}2008]{milne2008learning}
D.~Milne and I.~H. Witten.
\newblock 2008.
\newblock Learning to link with {Wikipedia}.
\newblock In {\em Proceedings of the International Conference on Information
  and Knowledge Management ({CIKM})}.

\bibitem[\protect\citename{Owoputi \bgroup et al.\egroup
  }2013]{owoputi2013improved}
Olutobi Owoputi, Brendan O'Connor, Chris Dyer, Kevin Gimpel, Nathan Schneider,
  and Noah~A. Smith.
\newblock 2013.
\newblock Improved part-of-speech tagging for online conversational text with
  word clusters.
\newblock In {\em {Proceedings of the North American Chapter of the Association
  for Computational Linguistics (NAACL)}}, pages 380--390, Atlanta, GA.

\bibitem[\protect\citename{Puniyani \bgroup et al.\egroup
  }2010]{puniyani2010social}
Kriti Puniyani, Jacob Eisenstein, Shay Cohen, and Eric~P. Xing.
\newblock 2010.
\newblock Social links from latent topics in microblogs.
\newblock In {\em Proceedings of {NAACL} Workshop on Social Media}, Los
  Angeles.

\bibitem[\protect\citename{Ritter \bgroup et al.\egroup }2011]{ritter2011named}
Alan Ritter, Sam Clark, Mausam, and Oren Etzioni.
\newblock 2011.
\newblock Named entity recognition in tweets: an experimental study.
\newblock In {\em {Proceedings of Empirical Methods for Natural Language
  Processing (EMNLP)}}.

\bibitem[\protect\citename{Shen \bgroup et al.\egroup }2013]{shen2013linking}
Wei Shen, Jianyong Wang, Ping Luo, and Min Wang.
\newblock 2013.
\newblock Linking named entities in tweets with knowledge base via user
  interest modeling.
\newblock In {\em {Proceedings of Knowledge Discovery and Data Mining (KDD)}}.

\bibitem[\protect\citename{Socher \bgroup et al.\egroup
  }2013]{SocherChenManningNg2013}
Richard Socher, Danqi Chen, Christopher~D. Manning, and Andrew~Y. Ng.
\newblock 2013.
\newblock {Reasoning With Neural Tensor Networks For Knowledge Base
  Completion}.
\newblock In {\em {Neural Information Processing Systems (NIPS)}}, Lake Tahoe.

\bibitem[\protect\citename{Speriosu \bgroup et al.\egroup
  }2011]{speriosu2011twitter}
Michael Speriosu, Nikita Sudan, Sid Upadhyay, and Jason Baldridge.
\newblock 2011.
\newblock Twitter polarity classification with label propagation over lexical
  links and the follower graph.
\newblock In {\em {Proceedings of Empirical Methods for Natural Language
  Processing (EMNLP)}}, pages 53--63.

\bibitem[\protect\citename{Tan \bgroup et al.\egroup }2011]{tan2011user}
Chenhao Tan, Lillian Lee, Jie Tang, Long Jiang, Ming Zhou, and Ping Li.
\newblock 2011.
\newblock User-level sentiment analysis incorporating social networks.
\newblock In {\em {Proceedings of Knowledge Discovery and Data Mining (KDD)}}.

\bibitem[\protect\citename{Tang \bgroup et al.\egroup }2015]{tang2015line}
Jian Tang, Meng Qu, Mingzhe Wang, Ming Zhang, Jun Yan, and Qiaozhu Mei.
\newblock 2015.
\newblock Line: Large-scale information network embedding.
\newblock In {\em {Proceedings of the Conference on World-Wide Web (WWW)}}.

\bibitem[\protect\citename{Wu \bgroup et al.\egroup }2011]{wu2011says}
Shaomei Wu, Jake~M Hofman, Winter~A Mason, and Duncan~J Watts.
\newblock 2011.
\newblock Who says what to whom on twitter.
\newblock In {\em {Proceedings of the Conference on World-Wide Web (WWW)}},
  pages 705--714.

\bibitem[\protect\citename{Yang and Chang}2015]{yang2015smart}
Yi~Yang and Ming-Wei Chang.
\newblock 2015.
\newblock S-mart: Novel tree-based structured learning algorithms applied to
  tweet entity linking.
\newblock In {\em {Proceedings of the Association for Computational Linguistics
  (ACL)}}, Beijing, China.

\bibitem[\protect\citename{Yang and Eisenstein}2015]{yang2015putting}
Yi~Yang and Jacob Eisenstein.
\newblock 2015.
\newblock Putting things in context: Community-specific embedding projections
  for sentiment analysis.
\newblock {\em arXiv preprint arXiv:1511.06052}.

\bibitem[\protect\citename{Yang \bgroup et al.\egroup }2014]{yang2014embedding}
Bishan Yang, Wen-tau Yih, Xiaodong He, Jianfeng Gao, and Li~Deng.
\newblock 2014.
\newblock Embedding entities and relations for learning and inference in
  knowledge bases.
\newblock {\em arXiv preprint arXiv:1412.6575}.

\bibitem[\protect\citename{Yih \bgroup et al.\egroup }2015]{yih2015semantic}
Wen-tau Yih, Ming-Wei Chang, Xiaodong He, and Jianfeng Gao.
\newblock 2015.
\newblock Semantic parsing via staged query graph generation: Question
  answering with knowledge base.
\newblock In {\em {Proceedings of the Association for Computational Linguistics
  (ACL)}}, Beijing, China.

\end{thebibliography}
